\pdfoutput=1
\documentclass[aps,pre,superscriptaddress,onecolumn,nofootinbib]{revtex4-2}
\usepackage{multirow} 
\usepackage{graphicx} 
\usepackage{amsmath,amssymb} 
\usepackage{hyperref} 

\graphicspath{ {./images/} }

\newcommand{\ourdata}{ZooLake } 
\newcommand{\bestav}{\textit{Best\_6\_avg} } 
\newcommand{\beststack}{\textit{Best\_6\_stack} } 

\begin{document}
\title[Lake Zooplankton Classification]{Deep Learning Classification of Lake Zooplankton} 
\author{S. Kyathanahally}
\email{sreenath.kyathanahally@eawag.ch}
\affiliation{Eawag, \"Uberlandstrasse 133, CH-8600 D\"ubendorf, Switzerland}
\author{T. Hardeman}
\affiliation{Eawag, \"Uberlandstrasse 133, CH-8600 D\"ubendorf, Switzerland}
\author{E. Merz}
\affiliation{Eawag, \"Uberlandstrasse 133, CH-8600 D\"ubendorf, Switzerland}
\author{T. Kozakiewicz}
\affiliation{Eawag, \"Uberlandstrasse 133, CH-8600 D\"ubendorf, Switzerland}
\author{M. Reyes}
\affiliation{Eawag, \"Uberlandstrasse 133, CH-8600 D\"ubendorf, Switzerland}
\author{P. Isles}
\affiliation{Eawag, \"Uberlandstrasse 133, CH-8600 D\"ubendorf, Switzerland}
\author{F. Pomati}
\affiliation{Eawag, \"Uberlandstrasse 133, CH-8600 D\"ubendorf, Switzerland}
\author{M. Baity-Jesi}
\email{marco.baityjesi@eawag.ch}
\affiliation{Eawag, \"Uberlandstrasse 133, CH-8600 D\"ubendorf, Switzerland}

\date{\today}

\begin{abstract}
Plankton are effective indicators of environmental change and ecosystem health in freshwater habitats, but collection of plankton data using manual microscopic methods is extremely labor-intensive and expensive. Automated plankton imaging offers a promising way forward to monitor plankton communities with high frequency and accuracy in real-time. Yet, manual annotation of millions of images proposes a serious challenge to taxonomists. Deep learning classifiers have been successfully applied in various fields and provided encouraging results when used to categorize marine plankton images. 
Here, we present a set of deep learning models developed for the identification of lake plankton, and study several strategies to obtain optimal performances, which lead to operational prescriptions for users. To this aim, we annotated into 35 classes over 17900 images of zooplankton and large phytoplankton colonies, detected in Lake Greifensee (Switzerland) with the Dual Scripps Plankton Camera. Our best models were based on transfer learning and ensembling, which classified plankton images with 98\% accuracy and 93\% F1 score. When tested on freely available plankton datasets produced by other automated imaging tools (ZooScan, FlowCytobot and ISIIS), our models performed better than previously used models. 
Our annotated data, code and classification models are freely available online.
\end{abstract}

\maketitle

\section{Introduction}

Plankton are a key component of the Earth’s biosphere. They include all the aquatic organisms that drift along with the currents, from tiny bacteria and microalgae, to larvae of vertebrates and invertebrates. Photosynthetic phytoplankton are responsible for about half of the global primary production~\cite{behrenfeld:01} and therefore play a central role in atmospheric carbon fixation and oxygen production. Zooplankton are a broad group of aquatic microorganisms, spanning over tens of thousands of species~\cite{sourina:91}, and comprising both carnivores and herbivores, the latter feeding on phytoplankton. Plankton are a critical component of aquatic food-webs, producing organic matter that forms the ultimate source of mass and energy for higher trophic levels~\cite{lotze:19}, and serve as food for fish larvae~\cite{banse:95}. The death and excretion of planktonic organisms results in massive amounts of carbon being sequestered, regulating the biological carbon pump locally and globally~\cite{volk:85}. Plankton biodiversity and dynamics therefore directly influence climate, fisheries {and the sustenance of} human populations {near water bodies}.

Planktonic organisms, being mostly small in size, have short lifespans and a strong sensitivity to environmental conditions, which makes their diversity and abundances very effective indicators of environmental change and ecosystem health, particularly in freshwater ecosystems that suffer from combined exposure to human local impacts and global change, such as warming and invasive species~\cite{williamson:09}. Information on individual plankton species is also critically important for the monitoring of harmful algal blooms, which can cause huge ecological and economical damage and have severe public health consequences~\cite{huisman:18}. The diversity and abundance of plankton is generally measured using labour intensive sampling and microscopy, which suffer from a number of limitations, such as high costs, specialised personnel, low throughput, high sample processing time, subjectivity of classification and low traceability and reproducibility of data. These limitations have stimulated the development of a multitude of alternative and automated plankton monitoring tools~\cite{lombard:19}, some of which were recently applied in freshwater systems~\cite{spanbauer:20,merz:21,tapics:21}. 

If, on one side, studying freshwater environments offers the opportunity to approach several issues related to i) automated recognition of plankton taxa in systems that are heavily monitored for water quality, and ii) the creation of plankton population time series useful for both research and lake management, on the other side it presents a series of practical advantages. 
The number of species present in a lake is in the order of few hundreds and community composition changes at the scale of decades ~\cite{pomati:12}, and virtually all lakes of the same region tend to share the same geographic/climatic region and the same species pool of plankton taxa~\cite{monchamp:19}. 
This would allow us to process real lake data with a diminished need to account for species variability, build rather quickly a database that comprises all seen taxa, and easily use our models for more than one site. 
Moreover, lakes are usually characterised by lower levels of non-planktonic suspended solids (e.g. sand, debris) compared to coastal marine environments, so one can expect to work with cleaner images, with a relatively small number of non-biological or non-recognizable objects being detected.

Among automated plankton monitoring approaches, imaging techniques have the highest potential to yield standardised and reproducible quantification of abundance, biomass, diversity and morphology of plankton across scales~\cite{lombard:19,merz:21}. Currently, several \textit{in-situ} digital imaging devices exists such as, FlowCytobot~\cite{olson:07}, Scripps Plankton Camera (SPC)~\cite{orenstein:20}, Video Plankton Recorder~\cite{davis:92}, SIPPER and a dual-magnification modified SPC (www.aquascope.ch)~\cite{merz:21}.

These digital imaging systems can produce very large volumes of plankton images, especially if deployed \textit{in-situ} for automated continuous monitoring~\cite{orenstein:20,merz:21}. While the extraction of image features that describe important plankton traits like size and shape are well established~\cite{orenstein:20,merz:21}, classifying large volumes of objects into different plankton taxonomic categories is still an ongoing challenge, and represents the most important component for plankton monitoring~\cite{macleod:10}. Automated classification of imaged plankton objects may substitute the 
time-consuming job of taxonomists~\cite{macleod:10} and allow sampling and counting taxa at high {temporal} and spatial resolution. Automation of plankton monitoring could represent a key innovation in the assessment and management of water quality, aquatic biodiversity, invasive species affecting ecosystem services (e.g. parasites, invasive mussels), and early warning for harmful algal blooms.

Automated plankton classification is characterized by a set of features that make this task less straightforward than other similar problems.
The data sets used for training, as well as the images analyzed after deployment, cover wide taxonomic ranges that are very unevenly distributed (some taxa are very common and others are rarely seen - this is called \textit{data imbalance} or \textit{class imbalance})~\cite{orenstein:15}, and this distribution changes over time, with new taxa appearing from time to time~\cite{schroder:18}. 
Moreover, many images do not belong to any taxon (\textit{e.g.} dirt), or they cannot be identified due to the low resolution, their position, focus, or being cropped.
Furthermore, labeling these data sets requires a high effort, because they need to be annotated by expert taxonomists, and sampling images from videos, as it is done \textit{e.g.} for camera traps~\cite{tabak:19}, is not helpful because the alignment of the organisms with respect to the camera does not generally change throughout the exposure time.

Image classification models fall into several broad categories, including unsupervised models (which clusters and classifies images without any manually-assigned tags), supervised models (which use a training library of manually identified images to develop the classification model), and hybrid models (which combine aspects of supervised and unsupervised learning).
Even though there is current research that relies on unsupervised learning~\cite{schroder:20,salvesen:20} or on the development of specific kinds of data preprocessing~\cite{zhao:10,zheng:17},
the current state of the art for classifying plankton data sets most often involves deep convolutional neural networks trained on manually classified images~\cite{py:16,dai:16,dai:17,lee:16,li:16,orenstein:17,cui:18b,luo:18,rodrigues:18,dunker:18,bochinski:19,lumini:19,kerr:20,lumini:20,eerola:20,henrichs:21,guo:21},\footnote{For a synthetic survey of relatively recent applications of deep and machine learning to plankton classification we refer the reader to Refs.~\cite{moniruzzaman:17,lumini:20}} which allow for a great flexibility across applications and were demonstrated more satisfactory than relying on the manual extraction of features~\cite{gonzalez:19}.
These applications very often resort to transfer learning~\cite{tan:18}, which consists of using models which were pretrained on a large image dataset (usually, ImageNet~\cite{deng:09}), and adapting them to the specific image recognition problem. 
Transfer learning was used in a two-step process to deal with data imbalance~\cite{lee:16}, but most commonly it is used because it allows for the training of very large models in reasonable times.
The main differences in the various applications to plankton often dwell in the kind of image preprocessing. For example,
Ref.~\cite{dai:17} filters the images in different ways, and feeds both the original and the filtered images as input to the models, Ref.~\cite{cui:18b} applies logarithmic image enhancement on black and white images, and Ref.~\cite{lumini:19} tests different ways of resizing the pictures.

Furthermore, several models can be used in synergy in order to obtain better performances (be it to deal with data imbalance or to reach a higher weighted accuracy). Two main approaches to combining multiple models are collaborative models and ensembling. The former consists of training models together to produce a common output~\cite{dai:17,kerr:20}, while the latter trains the models separately and combines the outputs in a later stage.
Collaborative models were used recently to counter data imbalance, yielding high performances on single-channel (\textit{i.e.} black and white) images obtained in Station L4 in the Western English Channel~\cite{kerr:20}. However, this involves deploying simultaneously several very heavy models, resulting in a very high memory usage, unless one uses smaller versions of the typically used models (thus, not allowing for transfer learning). Ensembling allows to fuse virtually any number of learners, and resulted in very satisfactory performances when joining different architectures (where DenseNets most often do best) or kinds of preprocessing~\cite{lumini:19}.\\
The mentioned methods for automated plankton classification were principally deployed in salt-water coastal habitats.
To our knowledge, the only previous work performing image classification on freshwater images is Ref.~\cite{hong:20}, where the data does not come from an automated system, and they study a small balanced dataset sorted in four categories (daphnia, calanoid, female cyclopoid, male cyclopoid), and obtain a maximum classification accuracy of 93\%.

In this paper, we study the classification of plankton organisms from lake ecosystems, on a novel dataset of lake plankton images that we make freely accessible, together with a code that allows to easily train and deploy our deep neural networks.
We analyze plankton images from the Dual-magnification Scripps Plankton Camera (DSPC), which is a dark field imaging microscope, currently deployed in lake Greifensee (Switzerland)~\cite{merz:21}, and specifically the images from the 0.5x magnification, which targets zooplankton and large colony-forming phytoplankton taxa in the ranges of $100~\mu$m to 1~cm. 
We manually annotated 17943 images consisting of $n_c=35$ unevenly distributed categories, which were collected \textit{in-situ} using the DSPC deployed at 3~m depth in lake Greifensee. 
We propose a set of latest deep learning models that makes use of transfer learning, and we combine them through versions of collaborative and ensemble learning. In particular we explore several ways to ensemble our models based on recent findings in statistics~\cite{geiger:20, dascoli:20}. 
We evaluate the performances of our models on publicly available datasets, obtaining a slight but systematic increase in performance with respect to the previous literature.
The simplest of the presented models were used to analyze part of the data in Ref.~\cite{merz:21}.

\section{Materials and Methods}\label{sec:methods}

\subsection{Data Acquisition}
\begin{figure}[tbh]
\centering
\includegraphics[width=\columnwidth]{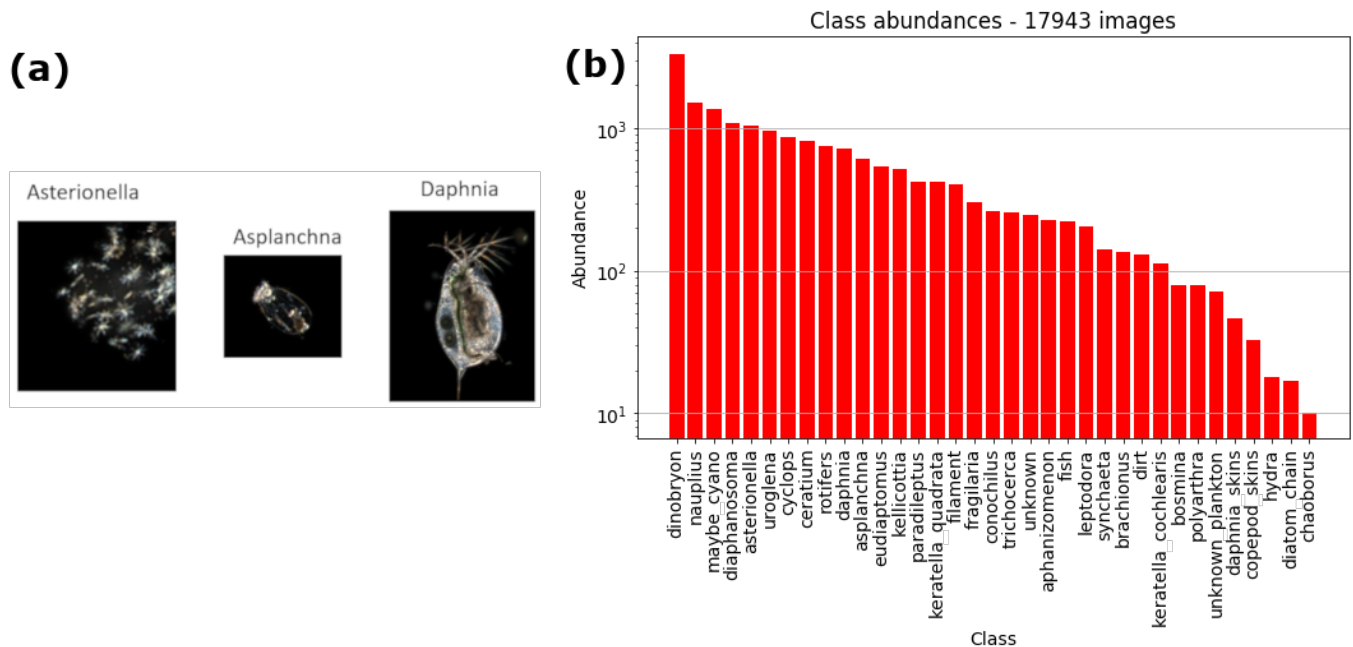}
\caption{\textbf{(a)}:~Sample images from the DSPC in lake Greifensee~\textbf{(b)}:~Abundance of each class in our dataset. The word \textit{class} is intended in the classification sense, and does not indicate the taxonomic rank. Note that the $y$ axis has a logarithmic scale.}
\label{fig:samples}
\end{figure}

We used images coming from the DSPC ~\cite{merz:21}, deployed in lake Greifensee, and acquired from wild plankton taxa across the years 2018 to 2020.\footnote{
Details on the camera and on the data acquisition can be found in Ref.~\cite{merz:21} (and Ref.~\cite{orenstein:20} for an analogous camera deployed in the ocean).}
The DSPC takes images of the microscopic plankton taxa  at user-defined frequencies and time intervals (for more details and camera settings see Ref.~\cite{merz:21}). The original full frame images may contain from zero to several images of planktonic organisms, as well as non-organic matter. The full frames are segmented on site in real time, and regions of interest (ROIs), which contain \textit{e.g.} plankton organisms, are saved and used for image feature extraction and classification. Images of objects at the boundary of the vision range of the camera result cropped, but we keep them anyway, as most of the time we are still able to identify them. The images have a black background, which favours the detection of ROIs. These have different sizes depending on the size of the detected object. For each ROI, we extracted 64 morphological and color features  
, and performed a series of graphical operations to make the image clearer.\footnote{
For details and code on image preparation we refer the reader to \href{https://github.com/tooploox/SPCConvert}{https://github.com/tooploox/SPCConvert}. This code contains the pipeline we used, of color conversion, edge-detection and segmentation, morphological feature extraction, foreground masking, and inverse filtering of masked foreground.
}
In Fig.~\ref{fig:samples}\textbf{(a)} we show some examples of what the final images look like.
In App.~\ref{app:dataset}, we provide an extensive description of the dataset and all its classes, together with one sample image from each class in Fig.~\ref{fig:all-class-images}.
In App.~\ref{app:features-decription} we describe the afore-mentioned 64 morphological features.

\subsection{Data Preparation}
The DSPC can be run with two different magnifications~\cite{merz:21}, but in this paper we report only on the images taken at the lower magnification, which contain mostly zooplankton taxa and several large colonial phytoplankton. We manually annotated a dataset of 17943 images of single objects, into $n_\mathrm{c}=35$ classes.\footnote{
Throughout this text, we use the machine-learning connotation of the work ``class", which indicates a category for classification, and not a taxonomic rank. In other words, our classes are not necessarily related to the taxonomic classification of the categories. For example, we call ``class" categories like ``diatom chain", ``unknown" or also ``dirt".
}
In Fig.~\ref{fig:samples}\textbf{(b)} we show the names of all the $n_c$ classes, along with the number of labeled images of each class.
Note that there are 300 times more annotated images of the most common class (dinobryon) than the rarest class (chaoborus).


\subsection{Open-access availability of our dataset}
We call \ourdata the described dataset of labeled plankton images.
We give extensive details on \ourdata in Apps.~\ref{app:dataset} and~\ref{app:features-decription},
and made the data openly available online
at the following link: \href{https://data.eawag.ch/dataset/deep-learning-classification-of-zooplankton-from-lakes}{https://data.eawag.ch/dataset/eep-learning-classification-of-zooplankton-from-lakes} .


\subsection{Further data preparation}
\label{sec:further-preparation}
Since for most deep learning models it is not convenient to have images of different sizes, we resized our images in such a way that they all had the same size. The two simplest ways of doing this are either by 
(i) resizing all the images to 128x128 pixels irrespective of its initial dimensions thus not maintaining the original proportions, or (ii) shrinking them in such a way that the largest dimension is at most 128 pixels (no shrinking is done if the image is already smaller) and padding them with a black background in order to make them 128x128.
The former method has the disadvantage of not maintaining proportions. The latter has the problem that in images with a very large aspect ratio there is a loss of information along the smallest dimension.\footnote{Imagine that an image is originally $1280\times50$ square pixels. Re-scaling the largest dimension to 128 pixels, maintaining the proportions, implies that there resulting image is only 5 pixels high, which means that we almost completely lose the information contained in the image.
Further, with method (ii), the large images are re-scaled, while the small ones are not, so even in this case the image size suffers a non-linear transformation.}
The two methods are compared in Ref.~\cite{lumini:19}, where it is seen that procedure (i) gives slightly better performances in most datasets. Further, the information lost when reshaping of the objects' aspect ratios can be recovered by using the initial aspect ratio (and similar quantities) as an extra input feature.
For these reasons, the results we show in the main text are all obtained through method (i).

In order to artificially increase the number of training images, we used data augmentation, consisting of applying random deformations to the training images.
The transformations we applied include rotations up to 180º, flipping, zooming up to 20\%, and shearing up to 10\%.
As for the morphological and color features, we calculated 44 additional ones (see App.~\ref{app:feature-models}), and standardized the resulting 111 features to have zero mean and unit standard deviation.

\subsection{Training, validation and test}
We split our images into training, validation and test sets, with a ratio of 70:15:15. The exact same splittings were used for all the models. 
The validation set was used to select the best model (hyper)parameters,
while the test set was set aside throughout the whole process, and used only at the very end to assess and compare the performance of all the proposed models.

\subsection{Deep learning models}
A common challenge when choosing deep learning architectures is how to best jointly scale  architecture depth, width and image resolution. A recent solution was given in Ref.~\cite{tan:19}, that proposes a scaling form for these three variables simultaneously, together with a baseline model, called EfficientNet-B0, for which this scaling is particularly efficient. This results in better performances than previous state of the art models, with a smaller investment in terms of model parameters and number of operations. The provided scaling form allows us to obtain efficiently scaled models according to how many computational resources we are willing to invest. These models, ordered by increasing size, are called Efficient-B1, Efficient-B2, Efficient-B3, Efficient-B4, Efficient-B5, Efficient-B6, and Efficient-B7.
Given the aforementioned large efforts to apply deep learning models to plankton classification, we believe that it is worth to assess the performances of these architectures on plankton recognition. Aside from those, we also test other deep neural network architectures, some of which were already used successfully for our kind of problems.

\begin{figure}[tbh]
    \centering
    \includegraphics[width=8cm]{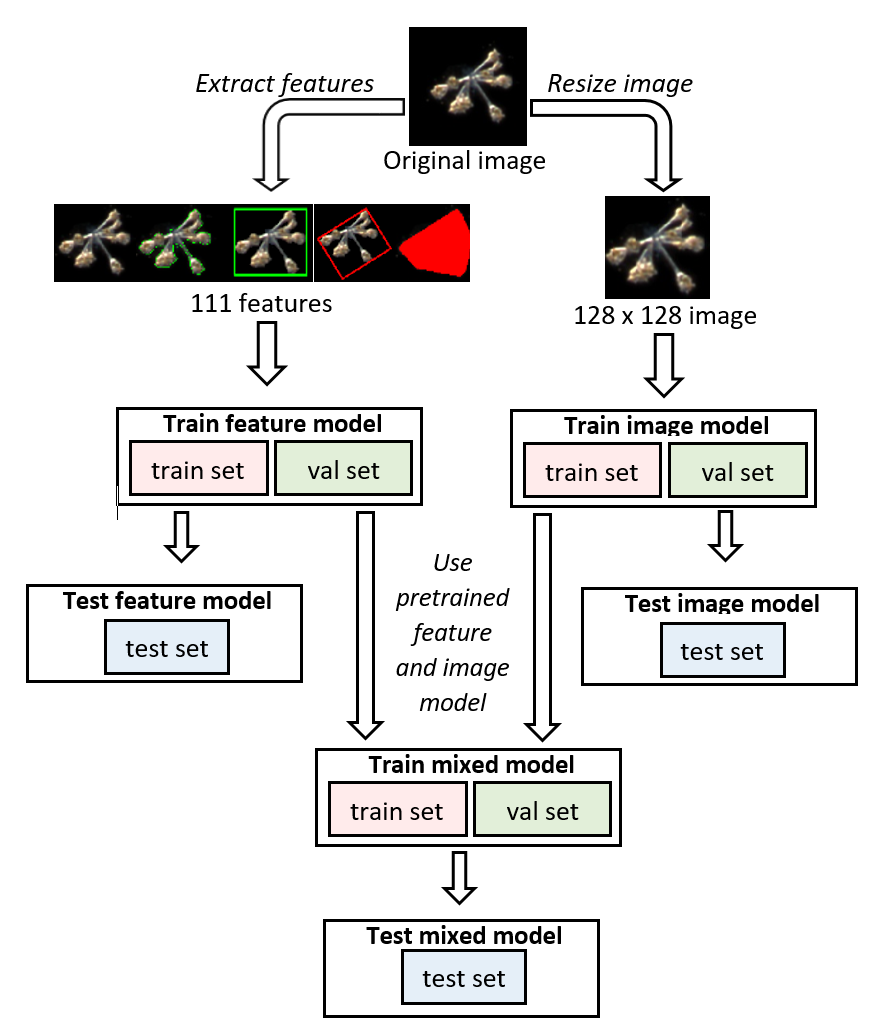}
    \caption{Diagram of the three main kinds of models that we mention in our paper. 
    Image models are convolutional networks that receive only images as input, feature models are MLPs that receive as input only features extracted from the image, but not the image \textit{per se}, and Mixed models join and fine-tune Image and Feature models.}
    \label{fig:pipeline}
\end{figure}

In the main text of this manuscript, we report on 12 different models. These are the EfficientNets B0 through B7~\cite{tan:19}, InceptionV3~\cite{szegedy:15}, Dense121~\cite{huang:18}, MobileNet~\cite{sandler:18} and ResNet50~\cite{he:15}), trained with transfer learning (Sec.~\ref{sec:transfer-learning}). 
Each individual model was trained four times, with different initial conditions from the same parameter distribution.\footnote{
All the initial conditions of all models were different realizations from the same distribution. We used a Glorot (or Xavier) uniform initializer, which is
a uniform distribution within $[-a, a]$, where $a = \sqrt{6 / (n_i + n_o)}$, and $n_i$ and $n_o$ are respectively the number of input and output units in the weight tensor.
All the models were trained with the Adam optimizer, 
a stochastic gradient descent method that is based on adaptive estimation of first-order and second-order moments. We used, respectively, 0.9 and 0.999 as decay rate of the first and second moment estimates.
} 
Additionally, we trained multi-layer perceptrons (MLPs) using as input the 110 morphological and color features mentioned in Sec.~\ref{sec:further-preparation}, and trained Mixed (collaborative) models that combine the MLPs with a larger model trained on images (Sec.~\ref{sec:mixed-models}). In Fig.~\ref{fig:pipeline} we sketch the structure of these Mixed models.
Finally, we also trained 4-layer convolutional networks, to assess whether through specific kinds of ensembling we could reach performances that match larger models (App.~\ref{app:ensembling}).\\

\subsection{Transfer learning}
\label{sec:transfer-learning}
Since training the mentioned models is a very demanding computational task, we used transfer learning, which consists of taking models that were already trained for image recognition on ImageNet, a very large dataset of non-planktonic images~\cite{russakovsky:15}.\footnote{Transfer learning from models trained on plankton images was tried in Ref.~\cite{orenstein:17}, but it did not yield better results than using the models trained on ImageNet.} 
We loaded the pretrained model and froze all the layers. We then removed the final layer, and replaced it with a dense layer with $n_c$ outputs, preceded and followed by dropout.
The new layers (dropout, dense, dropout, softmax with categorical cross-entropy loss) and learning rate were optimized with the help of the keras-tuner~\cite{omalley:19}.  We ran the keras-tuner with Bayesian optimization search,\footnote{The Bayesian optimization is a trial-and-error based scheme to find the optimal set of hyperparameters~\cite{mockus:12}.} 10 trials and 100 epochs, to find the best set of hyperparameters from the Bayesian search. Then, we trained for 200 epochs and used early stopping, \textit{i.e.} interrupting the training if the validation loss did not improve for 50 epochs, and keeping the model parameters with the lowest validation loss.
We then fine-tuned the model by unfreezing all the parameters and retraining again with a very low learning rate, $\eta=10^{-7}$, for 400 epochs. \\

\subsection{Ensemble learning}\label{sec:ensembling}
Ensemble methods use multiple independent learning algorithms to obtain better predictive performance than could be obtained from any of the constituent learning algorithms alone, often yielding higher overall classification metrics and model robustness~\cite{seni:10,zhang:12}. 
For our study we made use of two ensembling methods: averaging and stacking.\\

\subsubsection{Averaging}
For every image, the output of a single model is an $n_c$-dimensional confidence vector representing the probability that the model assigns to each class. The model's prediction is the class with the highest confidence.
When doing average ensembling over $n$ models, we take the average over the $n$ confidence vectors, and only afterwards choose the class with the highest confidence. With this procedure, all the models contribute equally to the final prediction, irrespective of their performance.
We performed average ensembling on the following choices of the models: 
\begin{enumerate}
    \item Across different models, as for example it was successfully done for plankton recognition in Refs.~\cite{lumini:19,lumini:20}.
    \item Across different instances of the same model, trained independently 4 times.
    This is inspired by the recent observation that this kind of averaging can lead to a better generalization in models with sufficiently many (but not too many) parameters~\cite{dascoli:20}. We provide a deeper discussion in App.~\ref{app:ensembling}.
    \item Manual selection of the six best individual models (on the validation set) over all the models. These best models resulted to be DenseNet121, EfficientNetB2, EfficientNetB5, EfficientNetB6, EfficientNetB7 and MobileNet. For each, we chose the initialization that gave the best validation performance.   
    We call this the \bestav ensemble model.
    \end{enumerate}

\subsubsection{Stacking}
Stacking is similar to averaging, but each model has a different weight. The weights are decided by creating a meta-dataset consisting of the confidence vectors of each model, and training a multinomial logistic regression on this metadataset.
We performed stacking both across initial conditions and across different architectures.
We call \beststack the ensemble model obtained by stacking the six individual best models (these are the same models that we used for the \bestav model).


\section{Results}
\subsection{Performances}
In Tab.~\ref{tab:single-model-performances} we summarize the performance of the individual models, along with the various forms of ensembling described in Sec.~\ref{sec:ensembling}. We report test accuracy and F1-score since, depending on the specific application, one can be interested in one metric or the other.
The accuracy is the fraction of correctly guessed images, so it is dominated by the most numerous classes. The F1-score is the geometric average between
precision and recall.\footnote{The recall (R) of class $i$ is the fraction of images belonging to class $i$ that were correctly labeled. The precision (P) of class $i$ is indicates how many of the examples that the classifier labeled as $i$ were correctly guessed. In terms of True Positives (TP), False Negatives (FN) and False Positives (FP), they are defined as
R=TP/(TP+FN), P=TP/(TP+FP), F1=2RP/(R+P).
}
We average the F1-score among the classes in such a way that
each class receives the same weight
regardless of how many images there were of that class.~\footnote{In other words, we calculate the macro-averaged F1-score.}

\begin{table*}[tbh]
\centering
\resizebox{\columnwidth}{!}{%
\begin{tabular}{ll|llll|ll}
\begin{tabular}[c]{@{}l@{}}Model \\ type\end{tabular} & \begin{tabular}[c]{@{}l@{}}Model \\ name\end{tabular} & \begin{tabular}[c]{@{}l@{}}Initial \\ condition 1 \\ (Accuracy/ \\ F1-score)\end{tabular} & \begin{tabular}[c]{@{}l@{}}Initial \\ condition 2 \\ (Accuracy/\\ F1-score)\end{tabular} & \begin{tabular}[c]{@{}l@{}}Initial \\ condition 3\\ (Accuracy/ \\ F1-score)\end{tabular} & \begin{tabular}[c]{@{}l@{}}Initial \\ condition 4 \\ (Accuracy/ \\ F1-score)\end{tabular} & \begin{tabular}[c]{@{}l@{}}Average \\ ensemble  \\ (Accuracy/\\ F1-score)\end{tabular} & \begin{tabular}[c]{@{}l@{}}Stacking \\ ensemble \\ (Accuracy/\\  F1-score)\end{tabular} \\ \hline \hline
Feature & MLP & 0.910/0.747 & {0.912/0.768} & 0.910/0.748 & 0.909/0.723 & 0.915/0.762 & 0.909/0.752 \\ \hline\hline
\multirow{12}{*}{Image} & Eff0 & 0.956/0.858 & 0.963/0.884 & {0.964/0.892} & 0.964/0.869 & 0.971/0.905 & 0.968/0.907 \\
 & Eff1 & 0.956/0.848 & 0.958/0.866 & {0.966/0.893} & 0.963/0.892 & 0.970/0.902 & 0.968/0.897 \\
 & Eff2 & \textit{0.967/0.893} & {0.967/0.899} & 0.968/0.894 & 0.966/0.889 & 0.975/0.915 & 0.969/0.913 \\
 & Eff3 & 0.958/0.841 & {0.957/0.880} & 0.959/0.877 & 0.958/0.868 & 0.969/0.904 & 0.965/0.883 \\
 & Eff4 & {0.958/0.876} & 0.964/0.870 & 0.962/0.874 & 0.962/0.873 & 0.972/0.903 & 0.970/0.907 \\
 & Eff5 & \textit{0.965/0.879} & {0.967/0.892} & 0.963/0.854 & 0.959/0.850 & 0.971/0.891 & 0.970/0.899 \\
 & Eff6 & 0.964/0.880 & \textit{0.965/0.879} & {0.968/0.897} & 0.964/0.865 & 0.971/0.904 & 0.970/0.912 \\
 & Eff7 & 0.966/0.885 & 0.970/0.899 & 0.967/0.886 & \textit{0.969/\textbf{0.900}} & 0.974/0.913 & 0.971/0.909 \\
 & IncepV3 & 0.965/0.876 & 0.961/0.883 & 0.954/0.867 & {0.964/0.884} & 0.972/0.901 & 0.971/0.913 \\
 & Dense121 & 0.958/0.859 & 0.962/0.821 & \textbf{0.971}/0.861 & \textit{0.968/0.890} & \textbf{0.976/0.916} & 0.975/0.884 \\
 & Mobile & 0.960/0.875 & \textit{0.959/0.891} & 0.958/0.886 & 0.965/0.870 & 0.971/0.907 & 0.971/0.907 \\
 & Res50 & {0.962/0.878} & 0.955/0.853 & 0.959/0.858 & 0.959/0.837 & 0.974/0.908 & 0.970/0.889 \\ \hline
\multirow{4}{*}{\begin{tabular}[c]{@{}l@{}}Image \\ Ensemble\end{tabular}} & Average & 0.976/0.911 & 0.977/0.923 & 0.975/0.909 & 0.976/0.914 & 0.977/0.919 &  \\ 
 & Stack & 0.975/0.908 & 0.976/0.919 & 0.976/0.914 & 0.977/0.915 &  & 0.978/0.921 \\
 & \bestav & \multicolumn{4}{c}{0.978/0.924} & \multicolumn{2}{l}{\multirow{2}{*}{}} \\
 & \beststack & \multicolumn{4}{c}{\textbf{0.979/0.927}} & \multicolumn{2}{l}{} \\ 
\end{tabular}}
\caption{Test accuracy and F1-score of the individual models across four different initial conditions. The rightmost and the bottom lines describe the performance of our ensemble models. The stacked model over all 48 image models performs stacking only once (we do not do two rounds of stacking). The entries in italics represent the six models that we chose for \bestav and \beststack based on the \textit{validation} F1-score (therefore their performance on the \textit{test} set is not necessarily the best).
In bold, we represent the overall best for each sector.
}
\label{tab:single-model-performances}
\end{table*}

We categorize the models in three ways, according to the kind of data they take as input. \textit{Feature models} take numerical features extracted from the images, \textit{image models} take the processed image, and \textit{mixed models} take both features and image.\\
\begin{figure*}[tbh]\centering
\includegraphics[width=14cm]{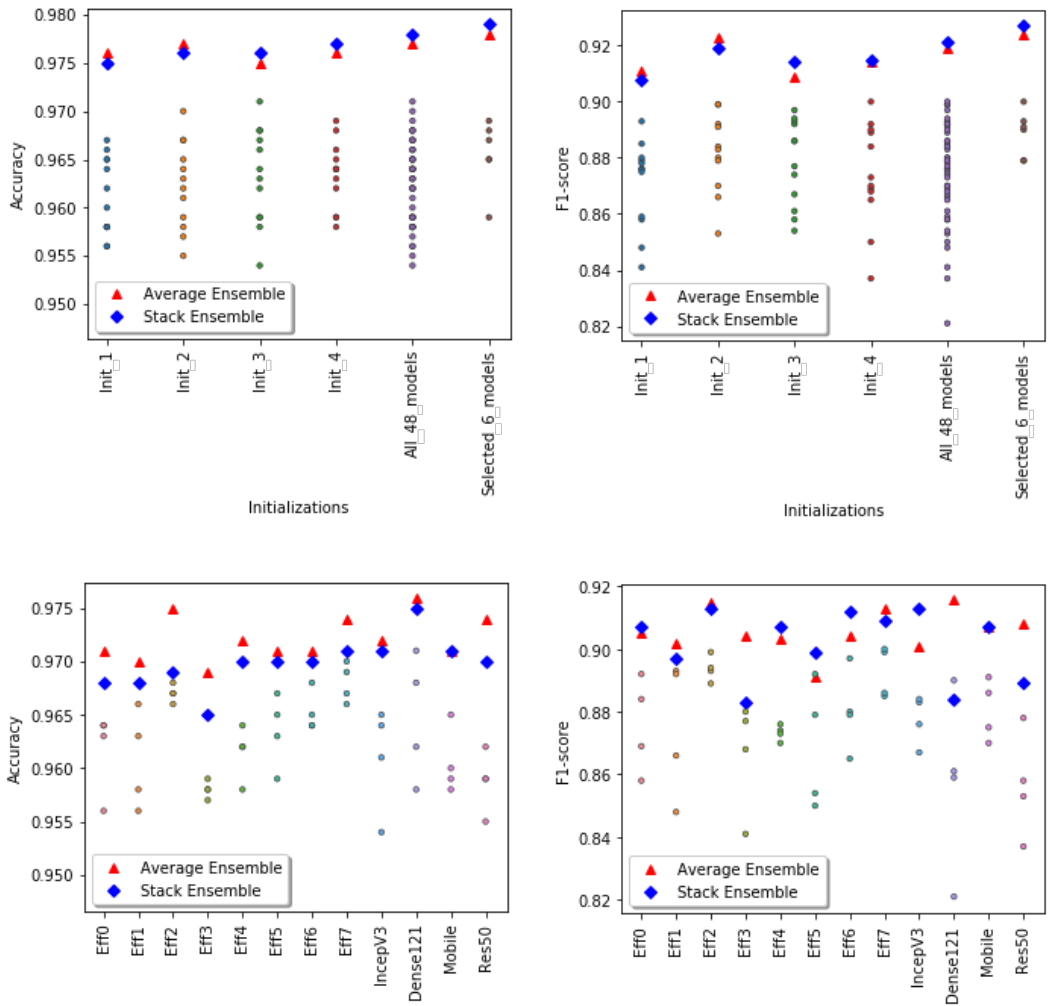}
\caption{Performance of single and ensemble models (same data as Tab.~\ref{tab:single-model-performances}). The solid circles are the single model performances.The red triangles represent the performance of average ensemble models, whereas the blue diamonds represent the performance of stacking ensemble models.
\textbf{Top-left}: The first four columns show the test accuracy of each initial condition across different models (this corresponds to going along a column in Tab.~\ref{tab:single-model-performances}). The fifth column shows the performances of all 48 image models, and the sixth is restricted to the models constituting \bestav and \beststack. In all cases we show the result of ensembling over these models.
\textbf{Top-right:} Same, but for the F1-score.
\textbf{Bottom-left:} For each model, we show the test accuracy of the four chosen initial conditions, and of ensembling through them (this corresponds to going along a row in Tab.~\ref{tab:single-model-performances}).
\textbf{Bottom-right:} Same, for the F1-score. The readers can refer to the table \ref{tab:single-model-performances} for the values of each single points on this figure
}
\label{fig:single-model-performances}
\end{figure*}

\subsubsection{Individual model performance}
First, we focus on the performances of the single models.
Already the MLP, our simplest model, which does not take the images as input, had a best accuracy of 91.2\%. 
However, the F1-score below 80\% reveals that the accuracy is driven by the predominant classes.\\
All the image models performed better than the MLP both in terms of accuracy and F1-Score. The model with the best F1-score is the EfficientNet-B7 (F1 = 90.0\%), followed by the EfficientNet-B2, which obtained almost the same value, but with a much smaller number of parameters ($8.4\times10^6$ parameters instead of $6.6\times10^7$ parameters for EfficientNetB7).\footnote{EfficientNet-B7 models also took about 8 hours to train, more than twice to train than their lightweight counterpart. We show the times required for hyperparameter tuning and for training in Tab.~\ref{tab:times}.} The lightest of the models we present is the MobileNet, with around $3.5\times10^6$ parameters, with a maximum F1-score of 89.1\%.

We tried to further improve the performance of EfficientNets by adopting basic methods for dealing with class imbalance. We reweighted the categories according to the number of examples of each class, in order to give an equal weight to all of them despite the class imbalance. We did not notice sizable improvements, so we restricted to only two models. We report on this in App.~\ref{app:imbalance}.\\[1ex]

\subsubsection{Ensembling across initial conditions}
As we discuss in App.~\ref{app:ensembling}, ensembling across initial conditions can help reduce the generalization gap (\textit{i.e.} the difference between train and test performance). This was shown for average ensembling~\cite{dascoli:20,geiger:20}, but we also tested it for stacking.
We see that (rightmost columns of Tab.~\ref{tab:single-model-performances}), both for stacking and averaging, this kind of ensembling improves the overall result compared to each individual model’s performance. We also show this in Fig.~\ref{fig:single-model-performances}--bottom, where in each column we show the performances of all the repetitions of a single model, as well as the result of ensembling through initial conditions. 
Average ensembling over (only four) initial conditions is very successful for some specific models such as Eff2 and Dense121.\\

\subsubsection{Ensembling across models}
We also ensembled across available models. For consistency, we first used only one initial condition per architecture  (randomly picked, without repetitions). The results shown in Tab.~\ref{tab:single-model-performances} and Fig.~\ref{fig:single-model-performances}--top (first four columns of each plot) display a clear improvement when performing this kind of ensembling, which in most cases seems more effective than over initial conditions.\\[1ex]

\subsubsection{Overall Ensembling}
Finally, we ensembled over all models and initial conditions, obtaining a further small improvement.
We obtained a slightly better improvement when ensembling on the six best models of the validation set (\bestav and \beststack), which had the further advantage of requiring less resources than using all 48 models. Our final best image model, \beststack, has an accuracy of 97.9\%, and an F1-score of 92.7\%. 

\begin{figure}[tbh]
\centering
\includegraphics[width=0.5\textwidth]{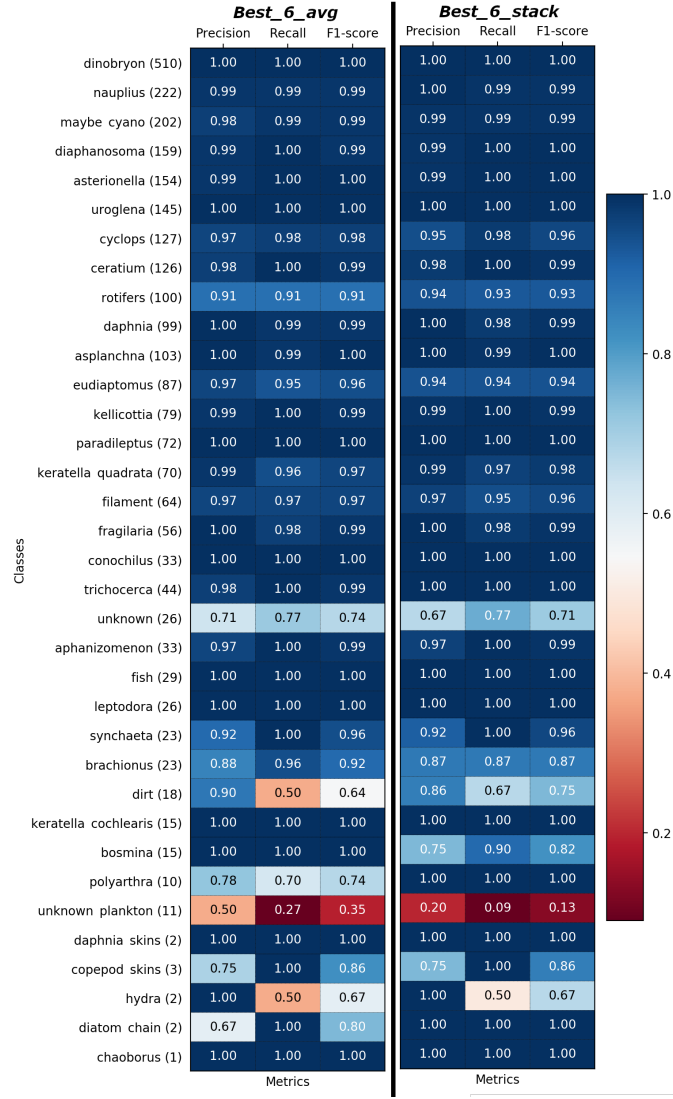}
\caption{Per-Class precision, recall and F1-score of \bestav (left) and {\beststack} (right) model on test set sorted based on Fig.~\ref{fig:samples}\textbf{(b)}.}
\label{fig:precision_recall}
\end{figure}

Towards practical purposes, the performances of \bestav and \beststack are even better than they appear if we take into account the nature of our dataset: the dataset is imbalanced, and for the most numerous two thirds of the classes we have almost perfect classification, as shown in Fig.~\ref{fig:precision_recall}, where we show the per-class performances. For the remaining third, the minority classes, the performance is good, though less reliable due to the very low number of test images at hand. 
If we keep into account the number of available images, the only three classes with a lower performance are the container (or junk) classes: unknown, dirt, unknown\_plankton.\footnote{We have a fourth container class, maybe\_cyano, but in that case we obtain almost perfect classification. See App.~\ref{app:dataset} for a description of this category.} 
This is not surprising, since these classes contain a wide variety of different objects, and it is less of a problem from the point of view of plankton monitoring, since misclassifications involving these classes are less relevant (we show the confusion matrices in Fig.~\ref{fig:conf_matrix}). 
If we exclude the three mentioned junk classes, we reach F1-score=97.3\%.
If we only consider the 23 classes for which the \ourdata dataset contains at least 200 examples (and keep the junk classes with $\geq200$ examples), the F1 scores go up to 98.0\%. 
Finally, if we both exclude the classes with less than 200 examples and the junk classes, we obtain F1-score=98.9\%.

Moreover, even when making mistakes, our models are not completely off. We can see this in Tab.~\ref{tab:top_k_acc}, where we plot the top-2 metrics of the \bestav and \beststack models. These represent how good the models' guesses are if the second choice of the classifier is considered as a success. We see that the macro-averaged recall increases by 3\%, and the total number of misclassified images is halved, with the top-2 accuracy exceeding 99\%.\\[1ex]
\begin{table}[tb]
    \centering
    \begin{tabular}{c||c|c}
         Model             & Macro Recall &  Accuracy\\\hline\hline
         \bestav (top-1)   &  0.926 &  0.978\\
         \bestav (top-2)   &  0.958 &  0.992\\\hline
         \beststack (top-1)&  0.928 &  0.979\\
         \beststack (top-2)&  0.947 &  0.988
    \end{tabular}
\caption{Top-1 and top-2 recall and accuracy. 
    Top-$n$ scores treat true positives and false negatives based on the $n$ highest values of the confidence vector. In other words, the top-2 scores are the model performances in the case that either of the top two guesses is correct.
    }    
    \label{tab:top_k_acc}
\end{table}

\subsection{Mixed models} \label{sec:mixed-models}
Since our image preprocessing did not conserve information on the image sizes, we trained mixed models that took as input a combination of image and 111 numerical features calculated from the image.

The numerical features were fed into the MLP described in App.~\ref{app:feature-models}, while the images were given as input to one of the image models described in Tab.~\ref{tab:single-model-performances}.
The two models were then combined and fed into a dense layer, followed by a softmax with categorical cross-entropy loss.

With both features and images (and no image augmentation) as input we trained with a low learning rate $\eta=10^{-5}$ for 400 epochs. For each choice of the initial conditions, each single image model was combined with its corresponding feature (MLP) model. In total, we trained 12 mixed models for 4 initial initial conditions each, so 48 mixed models in total.

Then, we ensembled through models and initial conditions in the same way as with the image models described in section \ref{sec:ensembling}. The test performance of the mixed models is shown in Table \ref{tab:mixed-model-performances}. 
The single-model performances are slightly better than those obtained through image-only models (Tab.~\ref{tab:single-model-performances}). However, after ensembling, the performance of mixed models becomes quite similar to that of image models.
The best F1 score of the mixed models improves that of the image models by 0.3\%, reaching 93.0\%.

\begin{table*}[tbh]
\centering
\resizebox{\columnwidth}{!}{%
\begin{tabular}{ll|llll|ll}
\begin{tabular}[c]{@{}l@{}}Model\\ type\end{tabular} & \begin{tabular}[c]{@{}l@{}}Model\\ name\end{tabular} & \begin{tabular}[c]{@{}l@{}}Initial\\ condition 1\\ (Accuracy/ \\ F1-score)\end{tabular} & \begin{tabular}[c]{@{}l@{}}Initial\\ condition 2\\ (Accuracy/\\ F1-score)\end{tabular} & \begin{tabular}[c]{@{}l@{}}Initial\\ condition 3\\ (Accuracy/\\ F1-score)\end{tabular} & \begin{tabular}[c]{@{}l@{}}Initial\\ condition 4\\ (Accuracy/\\ F1-score)\end{tabular} & \begin{tabular}[c]{@{}l@{}}Average\\ ensemble\\ (Accuracy/\\ F1-score)\end{tabular} & \begin{tabular}[c]{@{}l@{}}Stacking\\ ensemble\\ (Accuracy/\\ F1-score)\end{tabular} \\ \hline \hline
\multirow{12}{*}{Mixed} & Eff0+MLP & 0.962/0.874 & 0.969/0.857 & 0.968/0.867 & {0.966/0.882} & 0.973/0.917 & 0.963/0.856 \\
 & Eff1+MLP & 0.965/0.872 & 0.967/0.890 & {0.970/0.899} & 0.968/0.860 & 0.972/0.908 & 0.964/0.856 \\
 & Eff2+MLP & \textit{0.971/0.906} & 0.969/0.899 & {0.971/0.907} & 0.970/0.906 & \textbf{0.976/0.917} & 0.965/0.866 \\
 & Eff3+MLP & 0.964/0.864 & {0.965/0.904} & \textit{0.965/0.897} & 0.965/0.884 & 0.971/0.913 & 0.958/0.829 \\
 & Eff4+MLP & {0.967/0.897} & 0.968/0.864 & 0.967/0.884 & 0.968/0.886 & 0.973/0.909 & 0.962/0.847 \\
 & Eff5+MLP & {0.967/0.894} & 0.971/0.868 & 0.968/0.864 & 0.967/0.878 & 0.972/0.889 & 0.964/0.856 \\
 & Eff6+MLP & 0.971/0.881 & 0.971/0.891 & {0.971/0.897} & 0.967/0.873 & 0.974/0.914 & 0.966/0.863 \\
 & Eff7+MLP & 0.969/0.901 & \textit{\textbf{0.973/0.916}} & {0.973}/0.909 & 0.970/0.896 & 0.975/0.916 & 0.964/0.838 \\
 & IncepV3+MLP & 0.968/0.878 & \textit{0.965/0.893} & 0.962/0.888 & {0.970/0.896} & 0.973/0.911 & 0.965/0.842 \\
 & Dense121+MLP & 0.966/0.878 & 0.965/0.833 & 0.972/0.870 & \textit{0.972/0.881} & 0.974/0.881 & 0.962/0.836 \\
 & Mobile+MLP & 0.964/0.886 & \textit{0.966/0.899} & 0.962/0.893 & 0.970/0.879 & 0.971/0.904 & 0.964/0.857 \\
 & Res50+MLP & 0.965/0.861 & {0.964/0.890} & 0.963/0.857 & 0.965/0.856 & 0.971/0.875 & 0.964/0.856 \\ \hline
\multirow{4}{*}{\begin{tabular}[c]{@{}l@{}}Mixed\\ Ensemble\end{tabular}} & Average & 0.975/0.917 & 0.976/0.923 & 0.976/0.916 & 0.975/0.912 &  &  \\
 & Stack & 0.974/0.914 & 0.976/0.919 & 0.975/0.912 & 0.975/0.912 &  &  \\
 & {Best\_6\_avg} & \multicolumn{4}{c}{0.976/\textbf{0.930}} &  &  \\
 & {Best\_6\_stack} & \multicolumn{4}{c}{\textbf{0.977}/0.925} &  & 
\end{tabular}}
\caption{Mixed  model  test  accuracy  and  F1-score  of  the  individual models  across four different initial  conditions.   The trained  image  models  and  its  corresponding  feature  model  in  each  of  the  initial  conditions  were  chosen  from  Table   I.  The bottom four lines depict the performances when using the four kinds of ensembling described in the main text. The italics represent the six models that we chose for \bestav and \beststack based on the validation F1-score (therefore their performance on the test set is not always the best).
In bold, we represent the overall best for each sector.}
\label{tab:mixed-model-performances}
\end{table*}

\bigskip
\subsection{Comparisons with literature on public datasets of marine plankton images}

To compare our approach with previous literature, we evaluated our models on the publicly available datasets indicated in Ref.~\cite{zheng:17}, which reports classification benchmarks on subsets of the ZooScan~\cite{gorsky:10}, Kaggle~\cite{cowen:15}  and WHOI~\cite{orenstein:15} plankton datasets.
The ZooScan subset~\cite{zheng:17,gorsky:10} consists of 3771 greyscale images acquired using the Zooscan technology from the Bay of Villefranche-sur-mer. It consists of 20 classes with variable number of samples for each class. 
The Kaggle subset~\cite{zheng:17,cowen:15} comprises 14,374 greyscale images from 38 classes, acquired by In-situ Ichthyoplankton Imaging System (ISIIS) technology in the Straits of Florida and used for the National Data Science Bowl 2015 competition. The distribution among classes is not uniform, but each class has at least 100 samples.
The WHOI subset~\cite{zheng:17,orenstein:15} contains 6600 greyscale images of different sizes, that have been acquired by FlowCytobot~\cite{olson:07}, from Woods Hole Harbor water samples. The subset contains 22 manually categorized plankton classes with equal number of samples for each class. 

We compared the performance of our image models with the best models of Refs.~\cite{zheng:17,lumini:19,lumini:20}. For WHOI, we used the exact same train and test sets, since the dataset splitting was available.
For ZooScan and Kaggle we used respectively 2 fold cross-validation and 5 fold cross-validation as in Ref.~\cite{lumini:20}.
We used our \bestav and {\beststack} models, and did transfer learning starting from the weight configurations trained on our \ourdata dataset.\footnote{Since our \bestav and {\beststack} models were originally trained on three-channel image data, we had to adapt WHOI and Kaggle data images as they consisted of single channel images. The single channel was replicated 3 times to have 3 channels image such that they are similar to \ourdata. The ZooScan however had 3 channels images similar to \ourdata.}
We fine-tuned each of the 6 selected models belonging to \bestav and \beststack with a learning rate $\eta=10^{-5}$, and followed with average and stack ensembling.\footnote{We stress that for simplicity we used the 6 models that performed best on our ZooLake validation set. Arguably, we could expect an even higher performance if we selected the 6 models on the validation set of each of the three public datasets. We did not do this because it made the reporting more complicated, and our models perform better than the previous literature even in this case.} 

\begin{figure}[tbh]
\includegraphics[width=18cm]{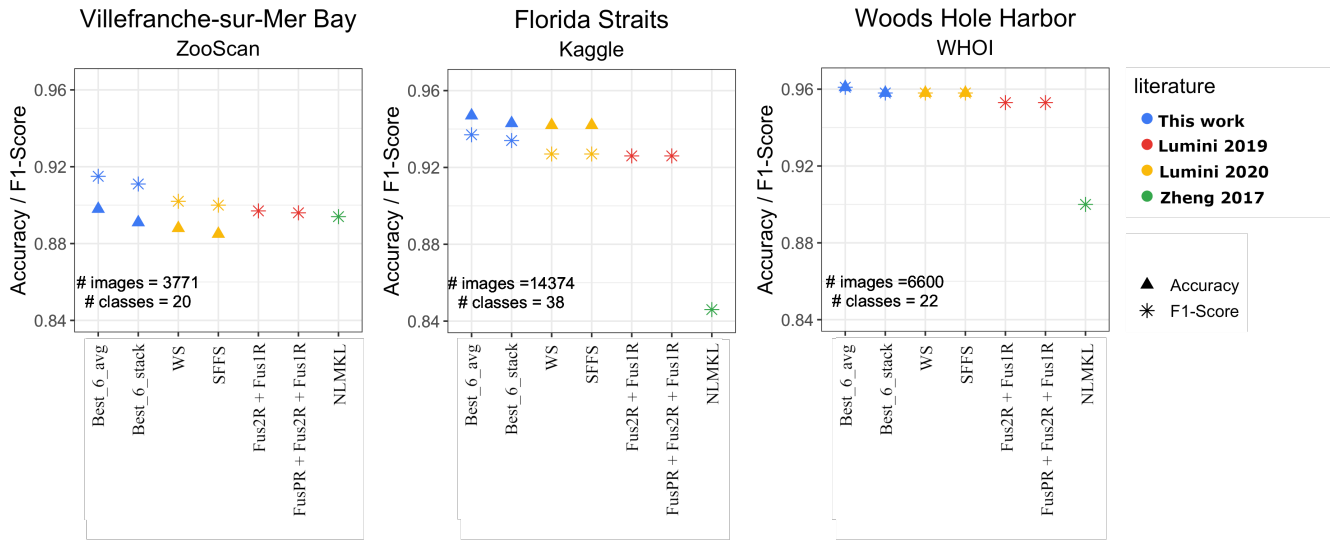}
\caption{Performances Accuracy/F1-score of our \bestav and \beststack models (blue points) on the publicly available datasets (ZooScan, Kaggle,  WHOI), and comparison with  previous  results from literature. 
The yellow points indicate ensemble models from Refs.~\cite{lumini:20}:
 SFFS (Sequential Forward Floating Selection - a feature selection method used to select models), WS (Weighed Selection - a stacking method that maximizes the performance while minimizing the number of classifiers). 
 The red points are the Fus models from Ref.~\cite{lumini:19}, which fuse diverse architectures and preprocessing. The green points stand for non-linear multi kernel learning (NLMKL), where an optimal non-linear combination of multiple kernels (Gaussian, Polynomial and Linear) is learnt to combine multiple extracted plankton features.}
\label{fig:whoi}
\end{figure}

As we show in Fig.~\ref{fig:whoi}, our \bestav and \beststack models performed always slightly better than all the previous methods/studies.
The improvement in terms of F1-score is consistent throughout the three datasets, with a 1.3\% improvement on the previously best model for ZooScan, a 1.0\% on Kaggle, and a 0.3\% on WHOI. The same data of Fig.~\ref{fig:whoi} is available in Tab.~\ref{tab:whoi}.\\
Note that these improvements come with a further advantage.
Our results require ensembling over a smaller number of models, and of total parameters. The 6-model average ensemble consisted of around $1.58\times10^8$ parameters compared to the $6.25\times10^8$ ($4.0$ times more) of the best model in Ref.~\cite{lumini:20} and the $1.36\times10^9$ ($8.6$ times more) of the best model in Ref.~\cite{lumini:19}. 
A major advantage of having lighter-weight models is that it allows for a simpler deployment and sharing with field scientists.


\section{Discussion}

In this paper, we presented the first dataset, to our knowledge, of lake plankton camera images, and showed that through an appropriate procedure of preprocessing and training of deep neural networks we can develop machine learning models that classify them with high reliability, reaching 97.9\% accuracy and 93.0\% (macro-averaged) F1-score. These metrics improve to 98.7\% accuracy  and 96.5\% F1-score if we exclude the few container classes (dirt, unknown, unknown\_plankton), that do not identify any specific taxon, with the F1 score reaching 98.9\% if we further restrict to the two thirds of the categories with a sufficient number of examples.
We made both the dataset and our code freely available.\footnote{Our \ourdata dataset can be downloaded at \href{https://data.eawag.ch/dataset/deep-learning-classification-of-zooplankton-from-lakes}{https://data.eawag.ch/dataset/deep-learning-classification-of-zooplankton-from-lakes}, and our code is available at \href{https://github.com/mbaityje/plankifier}{https://github.com/mbaityje/plankifier}.}

We trained several deep learning models. Our main novelties with respect to previous applications to plankton are the usage of EfficientNet models, a wise and simple ensemble model selection in the validation step, and the exploration of ensembling methods inspired by recent work in theory of machine learning~\cite{dascoli:20}.
We checked the utility of using mixed moedls that include as input numerical features such as the size of the detected object (in addition to the image), and found that this increases the single-model performance, but the gain is flattened out once we ensemble across several models (though the best F1 score still improved from 92.7\% to 93.0\%). 
We also checked whether the performance of the EfficientNets improved by correcting through class imbalance through class reweighting, and found no relevant improvement.
We compared the performances of our models with previous literature on salt-water datasets, obtaining an improvement that was steady across all datasets. 

The best performing individual models were EfficientNets, MobileNets and DenseNets. Surprisingly, the performance of the EfficientNets did not scale monotonously with the number of model parameters, perhaps due to the class imbalance of our dataset. The EfficientNets B2 and B7 were the best performing, but B2 uses a smaller number of parameters.
If we had to select a single architecture, our choice would lean towards MobileNet or EfficientNet-B2, given their favorable tradeoff between performance and model size.
If we apply ensembling, averaging and stacking provide similar performances, so we prefer averaging due to its higher simplicity. As for Mixed models, their narrow increase in performance after ensembling does not seem to justify their additional complexity in terms of deployment.

The Scripps Plankton Camera systems are a new technology that allows users to obtain large volumes of high-resolution color images, with virtually any temporal frequency. We noticed that the images that we obtained were clearer than those coming from marine environments (c.f. Ref.~\cite{orenstein:20}), which favoured the process of annotation and classification. 
Additionally, the taxonomic range is more stable during the seasonal progression compared to marine studies: fewer taxa are present in lake than coastal marine environments, colonisation by new taxa are relatively rare at the inter-annual scale (new taxa do not appear often), and lakes of the same region share large part of the plankton community composition. This makes the study of lake plankton dynamics an interesting and more controlled case study for method development due to its relative ecological simplicity and temporal stability, and implies that classifiers for lake taxa are more robust in these environments over space and time. This is particularly important from an application point of view, since the tools we developed in this paper are not only applicable for analyzing plankton population time series in lake Greifensee, addressing problems such as inferring interactions between taxa and predicting algal blooms, but they may be transferable to other similar lakes. Lakes represent very important water resources for human society and require routine monitoring for water quality and provision of ecosystem services. The models developed in this study can be directly used in real-world monitoring (\textit{e.g.} a preliminary version of our models was already used in Ref.~\cite{merz:21}).

\section*{Conflict of Interest Statement}

The authors declare that the research was conducted in the absence of any commercial or financial relationships that could be construed as a potential conflict of interest.

\section*{Author Contributions}
MBJ and FP designed the study. 
SK and MBJ built the models for Zooplankton classification.
TH, EW, TK, MR, PI, FP annotated the plankton images.
All the authors contributed to the manuscript.

\section*{Funding}
This project was funded by the Eawag DF project Big-Data Workflow (\#5221.00492.999.01), the Swiss Federal Office for the Environment (contract Nr Q392-1149) and the Swiss National Science Foundation (project 182124).

\section*{Acknowledgments}
We thank T. Lorimer and S. Dennis for contributions at initial stages of this study.

\section*{Data and Code Availability Statement}
We made both the dataset and our code freely available. Our \ourdata dataset, and the six models used for \bestav, can be downloaded at \href{https://doi.org/10.25678/0004DY}{\nolinkurl{https://doi.org/10.25678/0004DY}}. Our code is available at \href{https://github.com/mbaityje/plankifier/tree/For_publication}{\nolinkurl{https://github.com/mbaityje/plankifier/tree/For\_publication}}.


\appendix

\section{The \ourdata dataset}\label{app:dataset}
\subsection{Classes}
The \ourdata dataset contains 17943 images, sorted in 35 classes. The images were taken with the DSPC camera in lake Greifensee (Switzerland), between year 2018 and 2020.
Greifensee is a lake that we have been monitoring for many years and from which we know the plankton communities we can find. During the acquisition of the pictures, we have been sampling weekly for zoo and phytoplankton and identified the samples under a traditional microscope which has helped us learn to identify the pictures.

\begin{figure}[tbh]
\centering
  \includegraphics[width=11cm]{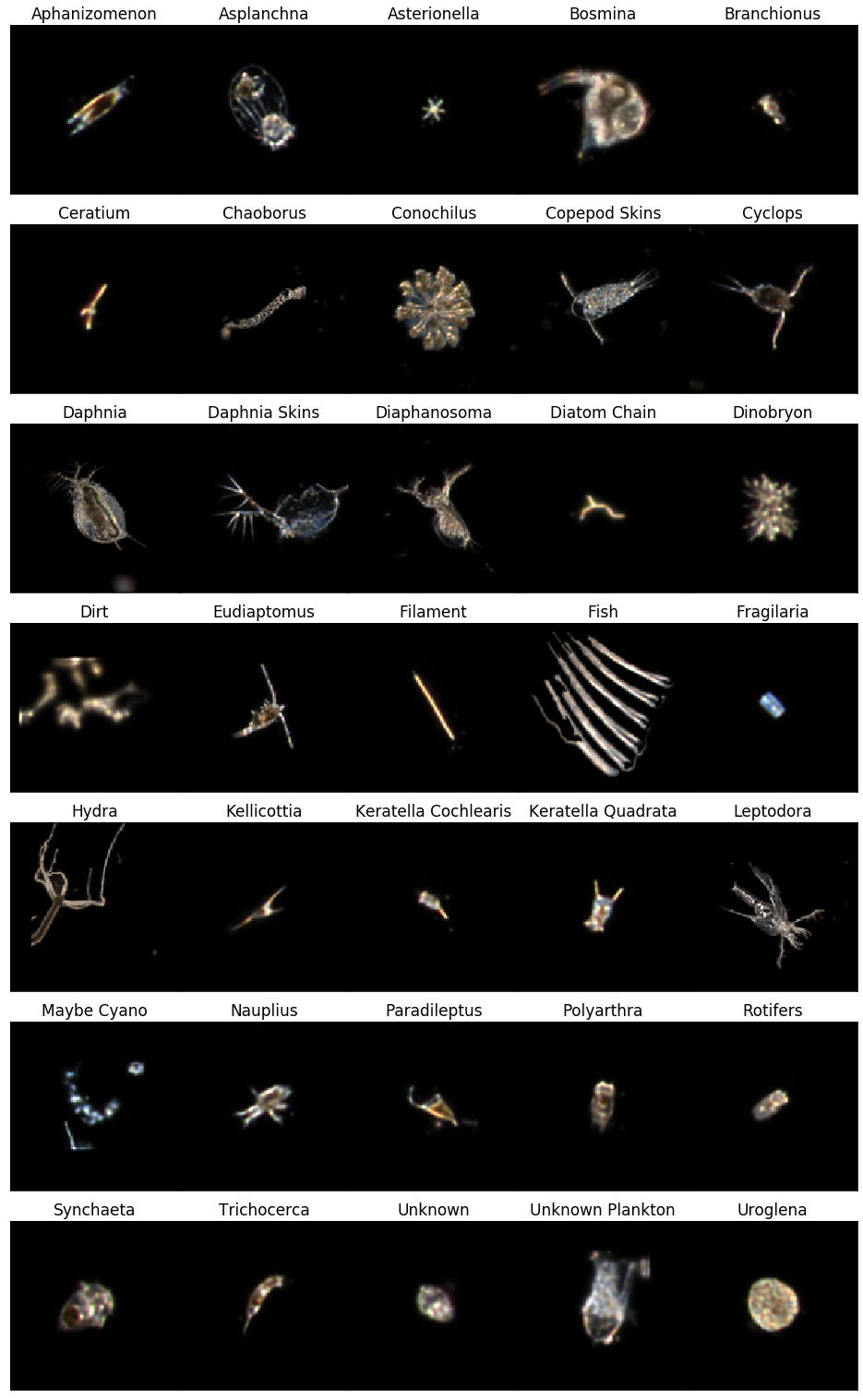}
  \caption{Examples of images from all 35 plankton classes in their original proportion.}
    \label{fig:all-class-images}
\end{figure}

Most of the classes identify specific plankton taxonomic categories. The low magnification camera used for this classifier was meant to categorize different zooplankton groups, but the high quality images, especially when sharp, has allowed us to be able to identify not only big zooplankton but also rotifers, and colony forming phytoplankton. We aimed at identifying the different categories to the maximum taxonomic resolution possible. 
In Fig.~\ref{fig:all-class-images} we show an example of image from each single class. 
In the following, we describe all the classes in the \ourdata dataset:
\begin{description}
    \item [aphanizomenon ] {These filamentous cyanobacteria are rarely solitary, mostly gathered in macroscopic fasciculated colonies, arranged in parallel. The single filaments would be almost impossible to identify to the genus level at the given magnification, but the colony form is very characteristic.}
    \item [asplanchna ] {Asplanchna is one of the biggest rotifers in our lake. It is a predator, semi transparent, shaped like a sack and with a well developed corona.}
    \item [asterionella ] {A diatom algae that builds stellate colonies. The colony itself is only 90um but it commonly makes aggregates forming the commonly known lake snow which are much bigger and hence seen in the low magnification camera. Interestingly when fixing the sample, following old traditional methods those aggregates are broken and one is not aware of the aggregates.}
    \item [bosmina ] {A zooplankton of the order Cladocera. Laterally compressed with almost spherical oblate body form and a characteristic long antennule.}
    \item [brachionus ] {Brachionus are rotifers with a sack shape body, and this is dorso-ventrally  flattened. The foot is long and not segmented}
    \item [ceratium ] {This category refers to the species Ceratium hirundinella. The distribution of the horns, size and color of this dinoflagellate makes it difficult to confuse with other species.}
    \item [chaoborus ] Pictures of the larvae of chaoborus, also commonly known as phantom midge. Chaoborus have the shape of a worm, mostly transparent and a distinguishable head capsule with Antennas.
    \item [conochilus ] {This rotifer forms spherical colonies that are radially oriented, attached in the center.}
    \item [copepod skins ] {We distinguished copepod pictures from copepods skins after molting (see also daphnia\_skins).}
    \item [cyclops ] {Zooplankton from the order of Cyclopoida, with antennas and 2 caudal appendages.}
    \item [daphnia ] {Zooplankton, a cladocera commonly known as water flea. Sometimes epiphytes or eggs were also clearly seen on the picture. This genus includes several species.}
    \item [daphnia skins ] {We know that Daphnia can molt, leaving the exoskeleton floating on the water several times during his life cycle. To avoid counting the exoskeleton as a Daphnia and hence increasing the real Daphnia concentration present in the water we distinguished this class from the daphnia category. The daphnia skins category pictures don't have any organels or eggs inside and are much more transparent.}
    \item [diaphanosoma ]{A genus of Cladoceras, zooplankton, similar to daphninds but with a peculiar thicker second Antennas.}
    \item [diatom chains ] {In this category we include few genuses of diatoms that make chains like filaments or zig zag. The magnification was not always enough to distinguish those colonies to a higher taxonomic level.}
    \item [dinobryon ] {The genus Dinobryon, a Chrysophyceae, is a colony forming algae that includes in this case at least 3 species. Cells in branching colonies, being D. bavaricum a bit more elongated than the other 2 species. Because the colonies are usually dense and big they can be seen in this magnification.}
    \item [dirt ] {In this category we included pictures of inorganic and organic material that were clearly not plankton.}
    \item [eudiaptomus ] {A zooplankton genus from the order calanoid. The first antennas are very long, more than half of the body. The eggsacks are very distinguishable from the ones from cyclops.}
    \item [filament ] { This folder includes different phytoplankton genus of a cylindrical shape and elongated. Appearing as a single filament.}
    \item [fish ] {This category includes all pictures were young fishes were partially photographed.}
    \item [fragilaria ] {This folder contains the colony forming diatoms \textit{Fragilaria crotonensis} and \textit{Fragilaria capucina}. they both have cells in large ribbon-like colonies. With the low magnification camera it was not possible to differentiate between this 2 species.} 
    \item [hydra ] {A macroscopic organism of the phylum cnidaria with a single body axis and tentacles. The body is a hollow tube with a gelatinous layer with a texture easy to identify with the camera.}
    \item [kellicottia ]{ The genus Kellicotia is a rotifer easy to identify with its elongated body and posterior and anterior spines.}
    \item [keratella cochlearis ] {This species of rotifer has an oval shape body with a long posterior spine and short anterior spines. It is sometimes difficult to distinguish from the brachionus rotifer, depending on the angle of the picture. }
    \item [keratella quadrata ] { In this category the taxonomic resolution of this rotifer is to species level because one can distinguish the 2 caudal spines at the base of the body.}
    \item [leptodora ] A top predator genus of zooplankton that can almost reach the size of 2cm. Most of the pictures of this category were of parts of its antennas or body, but still recognizable.
    \item [maybe cyano ] This algal category comprises different genuses of Cyanobacteria. All forming gelatinous colonies of different shapes with small cells inside. Due to the light and the darkfield background they look slightly blueish on the DSPC. This class possibly contains also non-organic material, like sand or debris, because they look very similar to some cyanobacteria colonies (especially clathrate microcystis colonies) and are hard to distinguish from one another with this magnification. The gathering of "maybe cyano" pictures started at blooms of cyanobacteria that were confirmed with the higher magnification camera, where cyanobacteria are more easily recognizable. Then, looking at pictures from the lower magnification camera (the one made for zooplankton and discussed in this paper), the taxonomists could be more sure about tagging the images they thought were cyanobacteria colonies. This is why the folder is called maybe cyano. With the 0p5x magnification it is very difficult to be sure; but with the expert knowledge, and seeing at the other higher magnification pictures taken at the same time, we could confirm that there were many cyanobacteria colonies at that time, which made us learn that cyanobacteria colonies pictures look very similar to the ones on this class, and enabled the tagging.
    \item [nauplius ] {In this zooplankton category we classified the larval stage of all copepods. Nauplii were distinguishable because of their antennas and mandible and an absence of thorax.}
    \item [paradileptus ] The genus paradileptus is a ciliate, has not been seen using traditional fixation methods with Lugol solution, that may had influence on its preservation~\cite{zarauz:08}. It has a conical body with a long tapering, spiraling neck region.
    \item [polyarthra ] { The most important feature of this rotifer is the well developed paddles (or blade-like projections) originating on the body below the head, and the absence of a foot, like in Brachionus}
    \item [rotifers ] {In this category we classified smaller rotifers, pictures of other rotifers that did not belong to any of the categories already containing rotifers: brachionus, conochilus, kellicotia, keratella, synchaeta or trichocerca, or pictures that were either not sharp or not from the right angle to be able to see the features to classify it into another category, since in some cases, in order to identify the rotifer to a genus level, one needs to see a trait that is only seen from a specific plane.}
    \item [synchaeta ] {Synchaetas are rotifers with a conical shape body that similarly to trichocerca have foot and toes, but this time those are reduced.}
    \item [trichocerca ] {The genus Trichocerca are rotifers with a lorica and a characteristic long 1 or 2 toes emerging from the foot. Anterior spines can be present but are not seen with the current magnification.}
    \item [unknown ] {Objects that were for us difficult to decide if they were dirt, or part of zooplankton or algae were ordered into an unknown category.}
    \item [unknown plankton ] {In this category we include all objects that we thought could be plankton but that because of the sharpness, focus or angle they were photographed we could not categorize them to a folder with label.}
    \item [uroglena ] {A Chrysophyceae algal genus with spherical colonies. They very often occur in blooms, so in very high densities.}
\end{description}


\subsection{Labeling}\label{app:labeling}
The dataset was labeled by a team of 6 taxonomists, listed among the authors of this paper. Every labeled image was checked by at least two taxonomists. As a further check,
we adopted a basic active learning procedure, in which the taxonomists occasionally double-checked the labels of the examples that were particularly poorly guessed by our classifiers.\\

\section{Features description}
\label{app:features-decription}
With the data, we include 64 features of quantities that are directly measured on the raw image.
Shape related features and color related features were extracted for every object in the image.\footnote{The code that we used to extract these features from the raw images is available at \href{https://github.com/tooploox/SPCConvert}{https://github.com/tooploox/SPCConvert}.}
The explanations of these features are given below.\\

\subsection{Shape related features:}

\begin{description}
    \item [aspect ratio] The ratio of width to height of the bounding rectangle of the object in the image 
    \item [eccentricity] This is the ratio of length of minor axis to the length of major axis of an object in the image
    \item [major and minor axis length] The longest perpendicular lines that can be drawn through the object in the image
    \item [orientation] The overall direction of the object in the image in degrees.
    \item [solidity] The ratio of the area of an object to the area of the convex hull of the object. This measures the density of an object.
    \item [estimated\_volume] This measures the estimated volume of the object in mm$^3$. 
    \item [area] This measures the area of the object in real world metric in mm$^2$.
\end{description}
    
\subsection{Color related features:}   
We calculated the following properties, for grayscale and colour components (R,G,B): 

\begin{description}
    \item [intensity\_mean] The mean average value of the intensity of image pixels for each R, G, B channel
    \item [intensity\_25\_percentile] The first quartile (25\%)  value of the intensity of  image pixels for each R, G, B channel
    \item [intensity\_50\_percentile] The second quartile (50\%) value of the intensity of image pixels for each R, G, B channel or in other words median value of the image pixels.
    \item [intensity\_75\_percentile] The third quartile (75\%) value of the intensity of image pixels for each R, G, B channel
    \item [intensity\_std] The standard deviation of the intensity of image pixels for each R, G, B channel - this shows how the pixels vary within the image
    \item [intensity\_mass\_displace] The distance between the peak intensity pixel value to the centre of the mass in an image for each R, G, B channel
    \item [intensity\_mass\_displace\_in\_images] The mass\_displace distance is scaled by the size of the input image for each R, G, B channel
    \item [intensity\_mass\_displace\_in\_minors] The mass\_displace distance is scaled by the minor axis length of the object in the image for each R, G, B channel
    \item [intensity\_moments\_hu] They give information on the shape and the intensity distribution of the image for each R, G, B channel. These Hu moment invariants are invariant to translation, scale and rotation.
\end{description}

\section{Feature models (multi-layer perceptrons)}\label{app:feature-models}

\begin{figure*}
\centering
  \includegraphics[width=15cm]{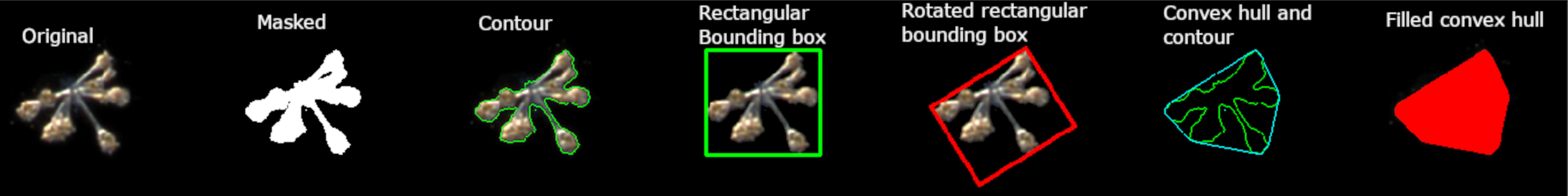}
  \caption{Illustrating some of the features extracted from the object in the image
}
  \label{fig:sample_extracted_features}
\end{figure*}

In addition to the 64 features of the \ourdata dataset, we extracted further features, in order to include some of those proposed in Ref.~\cite{kerr:20}. This implied adding 44 additional features.
These additional features are listed below. The reader can use Fig.~\ref{fig:sample_extracted_features} to understand some of the definitions.
\begin{description}
    \item [extent ]~is the ratio of contour area of the organism to bounding rectangle area.
    \item [equivalent Diameter ]~is the diameter of the circle whose area is same as the contour area of the organism.
    \item [image moments ]~helps to calculate the centroid, area, centre of the mass etc., of the object.
    \item [contour\_area and contour\_perimeter ]~is the area mm$^2$ and perimeter mm of the contour of the object.
    \item [hull\_area hull\_perimeter ]~is the area and perimeter of the convex hull. Convex hull is the smallest convex polygon that can fit in the object.
    \item [compactness ]~ is the contour\_perimeter squared, divided by $4\pi\times$contour\_area. The circle has a compactness of 1.
    \item [roundness ]~is similar to compactness, but here hull\_perimeter is used instead of contour\_perimeter i.e. $4\pi\times$contour\_area divided by hull\_perimeter squared. For a perfect circle, the roundness is 1.
    \item [w\_rot, h\_rot and angle\_rot ]~rotated width, height and angle of the rotated bounded rectangle.
    \item [rect\_width,rect\_height and rect\_area ]~width, height and area of the bounded rectangle over the object
    \item [rect\_area ]~width $\times$ height.
    \item [Convexity ]~hull\_perimeter/contour\_perimeter.
\end{description}

Using all 111 features as input, we trained multi-layer perceptron (MLP) models, a basic deep neural network model that alternates layers of matrix multiplication and non-linearities. We ran the Keras tuner~\cite{omalley:19} to get the optimized number of layers and model hyperparameters by setting Bayesian optimization search, 10 trials and 1000 epochs. As for the other models, this process involved only training and validation sets. Then, we trained for 200 epochs and with early stopping (with a patience of 50 epochs), and keeping the model parameters with the lowest validation loss. We concluded with a final 400 epochs of training with a learning rate $\eta=10^{-7}$.

The final feature model consisted of one input layer, one output layer and three hidden layers. The hidden layers consisted of 3 dense had each 0.3 dropout, and had respectively 128, 80 and 80 hidden units, and ReLU, tanh and softplus activations.

\section{Correcting for class imbalance} 
\label{app:imbalance}

Given the class imbalance of our dataset, we checked whether a basic method for dealing with data imbalance would improve the performances of our EfficientNet models.
We adopted class reweighting, giving minority classes more weight and viceversa~\cite{johnson:19}. During the training, misclassifications of minority classes are penalized more, and therefore the optimizer will be more keen to correct them.
We tested this on two models, EfficientNetB5 and EfficientNetB6, using exactly the same parameters (including those of the Bayesian search) as in \textit{Materials and Methods} section but including class reweighting. 
As shown in Tab.~\ref{tab:class-imbalance}, the F1-score did not improve. Therefore, class reweighting was not explored further. 
\begin{table}[tbh]
\centering
\begin{tabular}{lrr}
Model & Accuracy & F1-score \\ \hline \hline
Eff5 & 0.949 & 0.838 \\ 
Eff6 & 0.958 & 0.853 \\
\end{tabular}
\caption{Class-weightage technique’s performance on test data}
\label{tab:class-imbalance}
\end{table}

\section{Ensembling over initial conditions} \label{app:ensembling}
Recent work in Ref.~\cite{dascoli:20} on the double descent peak~\cite{belkin:19,nakkiran:19} showed that a major source of the generalization error is initialization variance. This variance can be attenuated by ensembling across different initializations of the same model. 
This was shown for simple balanced binary datasets in Refs.~\cite{dascoli:20,geiger:20}, and was especially useful near the interpolation threshold.\footnote{The interpolation threshold is the ratio between number of parameters and amount of data at which we are able to reach approximately zero error.}

In our case, we did not know where the interpolation threshold was, but we could assume that the models shown in Tab.~1 of the main text are highly over parameterized, and even then models such as Eff4 enjoyed over a 2\% improvement thanks to ensembling over only 4 initial conditions. 

\begin{figure*}[tbh]
\centering
\includegraphics[width=\columnwidth]{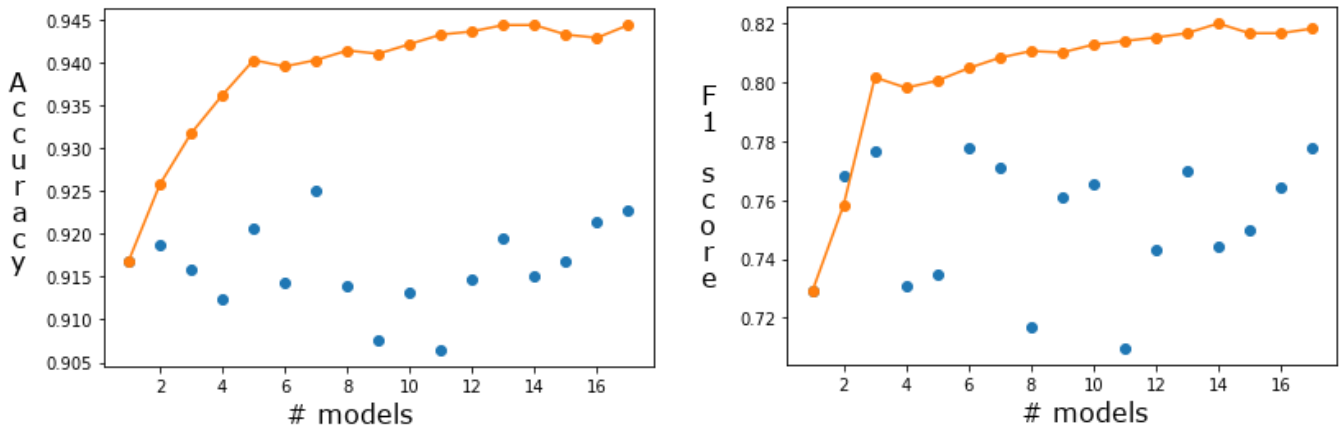}
\caption{Test accuracy (left) and F1-score (right) of four-layer convolutional networks. Blue dots represent the performance of individual model whereas the orange dots represent the cumulative performance of average ensembles of the model.}
\label{fig:conv4}
\end{figure*}

We can then expect that smaller convolutional networks (thus, closer to the interpolation threshold), can benefit more from ensembling, and we might be able to reach accuracies similar to those of EfficientNets with much simpler models.
Therefore, we trained convolutional networks that are close to the interpolation threshold, and investigated the improvement obtained through ensembling.
In Fig.~\ref{fig:conv4} we show results from a four-layer convolutional network (conv4) obtained through a Bayesian optimization search, 25 trials, 30 epochs to select the best convolutional network between 2 to 5 convolutional layers, filter size between 32 and 128, kernel size between 8 to 32, dense units between 64 to 256, learning rate between $10^{-2}$ and $10^{-5}$. We used no early stopping here, since ensembling should replace its efficacy~\cite{geiger:20}.
The average final training accuracy that we obtained was 99.24\%.

We ran the model with $K=17$ different initial conditions.
Fig.~\ref{fig:conv4} shows the test metrics while ensembling over an increasing number of initial conditions.
Blue circles represent the performance of individual conv4 models, whereas the orange dots represent the cumulative performance of average ensembles of the model. 
The benefit of ensembling is clearly visible, as already with five initial conditions we reach an accuracy (F1-score) around 0.94 (0.8). 
Ensembling over 17 initial conditions give a 0.945 accuracy (0.82 F1-score). 
This is likely to increase if we increase $K$, but it does not seem that the performances of the EfficientNets can be reached.

\section{Supplementary Tables and Figures}
\subsection{Figures}
\begin{figure*}[tbh]
  \includegraphics[width=15cm]{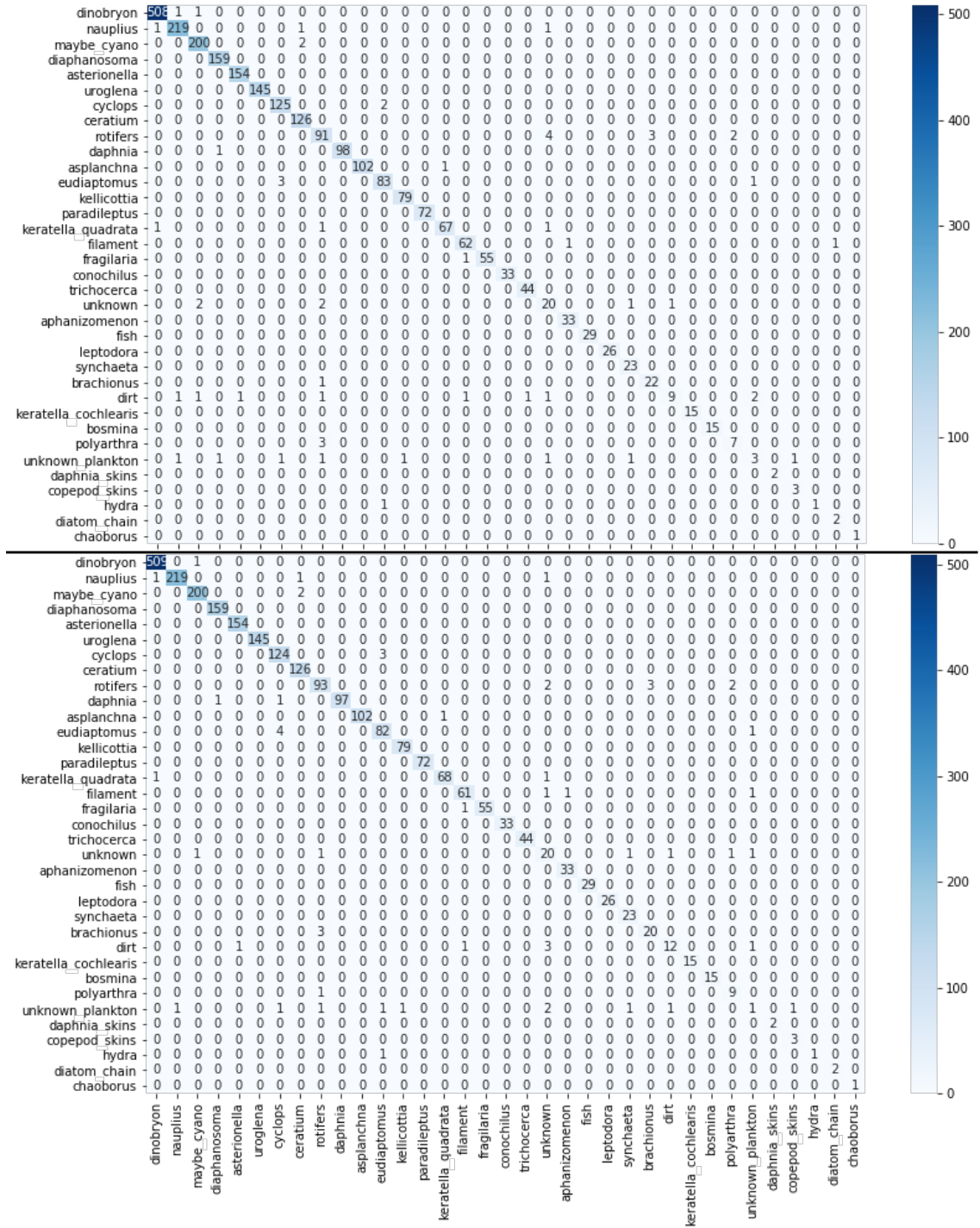}
  \caption{Confusion matrix of \bestav (top) and {\beststack} (bottom) Image models on test set sorted by overall abundance.}
\label{fig:conf_matrix}
\end{figure*}

In Fig.~\ref{fig:conf_matrix}, we show the confusion matrices of the \bestav and \beststack models.\\

\subsection{Tables}
In Tab.~\ref{tab:whoi} we compare the performance of our models on public datasets, with previous literature. The numbers in the table correspond to the data shown in Fig.~5 of the main text.
\begin{table*}[tbh]
\centering
\resizebox{\columnwidth}{!}{%
\begin{tabular}{l|ccc}
Model & ZooScan & Kaggle & WHOI \\ \hline \hline
\bestav (ours)                        & \textbf{0.898/0.915} & \textbf{0.947/0.937} & \textbf{0.961/0.961} \\ \hline
\beststack (ours)                     & 0.891/0.911 & 0.943/0.934 & 0.958/0.958 \\ \hline
SFFS~\cite{lumini:20} (11 classifiers)& 0.885/0.900 & 0.942/0.927 & 0.958/0.958 \\ \hline
WS~\cite{lumini:20} (11 classifiers)  & 0.888/0.902 & 0.942/0.927 & 0.958/0.958 \\ \hline
Fus\_2R+Fus\_1R~\cite{lumini:19} (24 classifiers) & n.a./0.897 & n.a./0.926 & n.a./0.953 \\ \hline
Fus\_PR+Fus\_2R +Fus\_1R~\cite{lumini:19} (43 classifiers) & n.a./0.896 & n.a./0.926 & n.a./0.953 \\ \hline
NLMKL~\cite{zheng:17} (3 kernels)       & n.a./0.894 & n.a./0.846 & n.a./0.900 \\ \hline
\end{tabular}}
\caption{Performances Accuracy/F1-score of our \bestav and \beststack models on the publicly available datasets (ZooScan, Kaggle,  WHOI), and comparison with  previous  results from literature. 
Different ways of creating ensembles are identified with the keywords SFFS (Sequential Forward Floating Selection - a feature selection method used to select models), WS (Weighed Selection - a stacking method that maximizes the performance while minimizing the number of classifiers) and Fus (fusion of diverse architectures and preprocessing) indicate different ways of creating ensembles~\cite{lumini:19,lumini:20}. The last line stands for non-linear multi kernel learning (NLMKL), where an optimal non-linear combination of multiple kernels (Gaussian, Polynomial and Linear) is learnt to combine multiple extracted plankton features.
}
\label{tab:whoi}
\end{table*}

In Tab.~\ref{tab:times} we show the amount of wallclock time required to train our models on an NVIDIA GeForce RTX 2080 Ti GPU.

\begin{table*}[tbh]
\centering
\resizebox{0.5\textwidth}{!}{%
\begin{tabular}{l|cc}
Model   & Training time [hours] &  Bayesian search [hours] \\ \hline \hline
Eff0    & 3.8 &  14.8 \\
Eff1    & 3.8 &  15.3 \\
Eff2    & 3.0 &  15.6 \\
Eff3    & 3.7 &  15.1 \\
Eff4    & 3.5 &  14.7 \\
Eff5    & 4.5 &  14.9 \\
Eff6    & 5.5 &  15.3 \\
Eff7    & 7.9 &  15.9 \\
Mobile  & 3.2 &  14.7 \\
Dense121& 3.7 &  15.0 \\
IncepV3 & 3.0 &  15.5 \\
Res50   & 3.2 &  15.0 \\
\end{tabular}}
\caption{Training times of our image models on an NVIDIA GeForce RTX 2080 Ti GPU. The second column represents the time to train a model across all phases of training, for a single choice of initial conditions and hyperparameters.
The third column depicts the time required to choose the best hyperparameters.
}
\label{tab:times}
\end{table*}

\bibliographystyle{unsrt} 
\bibliography{marco}

\end{document}